\documentclass{vivo}

\usepackage{booktabs} 
\usepackage{colortbl} 
\usepackage{multirow} 

\definecolor{bestresult}{RGB}{255, 232, 232}
\definecolor{secondresult}{RGB}{255, 247, 210}

\usepackage{graphicx}
\usepackage{wrapfig}
\usepackage{capt-of}

\newcommand{\quantitativeresultstable}{
\begin{table*}[!t]
  \caption{Quantitative comparison on the evaluation dataset. Red and yellow cells indicate the \colorbox{bestresult}{best} and \colorbox{secondresult}{second-best} results.}
  \label{tab:quantitative_results}
  \centering
  \resizebox{\textwidth}{!}{
  \begin{tabular}{lccccc}
    \toprule[1.1pt]
    \multirow{2}{*}{Method} & \multicolumn{3}{c}{\textbf{VBench}~\cite{huang2023vbench}} & \textbf{VLM-as-Judge} & \textbf{User Study} \\
    \cmidrule[0.8pt](lr){2-4}\cmidrule[0.8pt](lr){5-5}\cmidrule[0.8pt](lr){6-6}
    & Motion Smoothness $\uparrow$ & Temporal Flickering $\uparrow$ & Aesthetic Quality $\uparrow$ & Physical Similarity $\uparrow$ & Preference Rank $\downarrow$ \\
    \midrule[1.1pt]
    Wan2.1-I2V-14B & 97.86 & 96.34 & 50.23 & 2.67 & \multirow{3}{*}{2 (33.3\%)} \\
    Wan2.2-TI2V-5B & 98.03 & 96.97 & 49.20 & 2.61 & \\
    Wan2.2-I2V-14B & 98.58 & 97.10 & \cellcolor{secondresult}50.84 & \cellcolor{secondresult}2.97 & \\
    \midrule
    VAP-CogVideo & 98.58 & 97.44 & 50.19 & 2.92 & \multirow{2}{*}{3 (5.9\%)} \\
    VAP-Wan & 97.32 & 95.88 & 49.03 & 2.87 & \\
    \midrule
    VACE & \cellcolor{bestresult}98.73 & \cellcolor{bestresult}97.81 & 49.33 & 2.14 & 4 (2.0\%) \\
    \midrule
    \textbf{VIPER (Ours)} & \cellcolor{secondresult}\textbf{98.60} & \cellcolor{secondresult}\textbf{97.45} & \cellcolor{bestresult}\textbf{51.86} & \cellcolor{bestresult}\textbf{3.42} & \cellcolor{bestresult}\textbf{1 (58.8\%)} \\
    \bottomrule[1.1pt]
  \end{tabular}
  }
  \vspace{-2mm}
\end{table*}
}

\newcommand{\ablationtable}{
\begin{table}[htbp]
  \caption{Ablation study on the learnable queries and training strategy.}
  \label{tab:ablation}
  \centering
  \resizebox{\columnwidth}{!}{
  \begin{tabular}{lcccc}
    \toprule
    Variant & Motion S. $\uparrow$ & Temporal F. $\uparrow$ & Aesthetic Q. $\uparrow$ & Physical S. $\uparrow$ \\
    \midrule
    w/o learnable queries & 98.02 & 97.42 & 34.17 & 2.00 \\
    w/o Stage 1 & 98.51 & 97.08 & 51.01 & 3.41 \\
    w/o Stage 3 (LoRA) & 98.49 & 97.24 & 50.57 & 3.33 \\
    w/o Stage 1 \& Stage 2 & 98.45 & 97.03 & 51.05 & 3.26 \\
    \textbf{Full} & \cellcolor{bestresult}\textbf{98.60} & \cellcolor{bestresult}\textbf{97.45} & \cellcolor{bestresult}\textbf{51.86} & \cellcolor{bestresult}\textbf{3.42} \\
    \bottomrule
  \end{tabular}
  }
  \vspace{-2mm}
\end{table}
}

\usepackage[ruled]{algorithm2e} 

\SetAlFnt{\small}
\SetAlCapFnt{\small}
\SetAlCapNameFnt{\small}
\SetAlCapHSkip{0pt}

\begin{document}
\title{VIPER: Visual In-Context Physics Reasoning for Physically Plausible Video Generation}


\author[]{Tianxiao Chen$^{1,2}$} 
\author[]{Hanmo Chen$^{2}$}
\author[]{Huajin Chen$^{1}$}
\author[]{Bo Li$^{2}$}
\author[]{Qi Ye$^{1\dagger}$}
\author[]{Peng-Tao Jiang$^{2\dagger}$}

\affiliation[]{$^{1}$Zhejiang University} 
\affiliation[]{ \\ $^{2}$vivo BlueImage Lab, vivo Mobile Communication Co., Ltd.}

\affiliation[]{ \\ $\dagger$: Corresponding authors.}

\abstract{Modern video generation models can synthesize visually compelling and temporally coherent clips, yet controlling their physical behavior remains difficult with standard text and image conditions. The core challenge is a \textbf{conditioning bottleneck}: material response, contact interaction, deformation, and motion trajectory are continuous and relational physical cues that are hard to specify exhaustively in language but can be demonstrated naturally by video. We propose \textbf{VIPER}, a Visual In-Context Physics Reasoning framework for reference-guided image-to-video generation. Given a target image, a brief target prompt, and a reference video, VIPER treats the reference as a dense visual demonstration of the desired physical process rather than an appearance template. It uses a Multimodal Large Language Model (MLLM) to extract reference-derived physical cues and guide a pretrained image-to-video generator through a hierarchical training strategy, enabling physical behavior transfer while preserving the visual prior of the base generator. To support this setting, we construct \textbf{VIPER-19K}, a curated dataset with material, trajectory, and physical-impact annotations, together with filtered reference-target pairs. Experiments on an unseen validation set show that VIPER achieves stronger reference-video physical similarity and higher human preference than representative video generation and video-as-prompt baselines, while maintaining competitive general video quality. Qualitative results further demonstrate that VIPER can transfer reference-derived physical behavior to new target scenes without requiring carefully engineered prompts.
}

\maketitle

\begin{figure}[t]
  \centering
  \includegraphics[width=\textwidth]{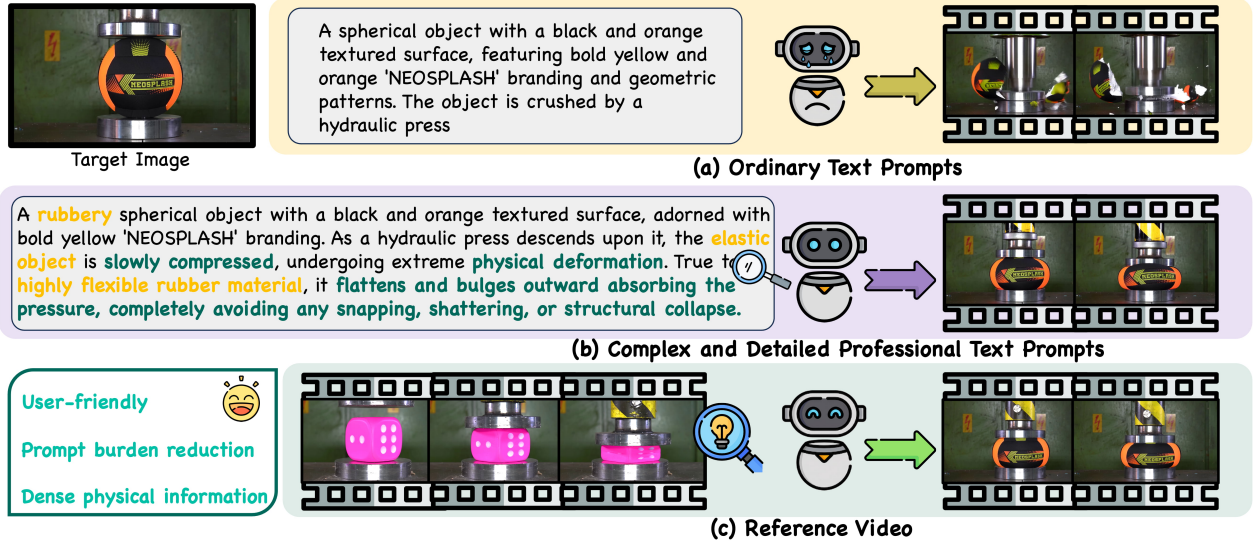}
  \caption{\textbf{Motivation}. (a) With an ordinary prompt, a video generator may produce visually plausible but physically implausible results because the physical process is under-specified. (b) A detailed and carefully engineered prompt can better communicate the expected physical behavior, but such prompts are difficult to write in practical use. (c) VIPER uses a reference video as a dense visual in-context demonstration, allowing the model to generate the target video with the expected physical behavior without requiring an exhaustive textual description.}
  \label{fig:teaser}
\end{figure}

\section{Introduction}
\label{sec:introduction}

Recent video generation models~\cite{sora,hunyuanvideo,wan2025wan} have rapidly improved in generating visually plausible and temporally coherent videos, benefiting from large-scale training data, increased model capacity, and advances in diffusion-transformer architectures~\cite{dit}.
Despite substantial progress in visual fidelity and temporal coherence, generating videos that follow real-world physical regularities remains challenging, as physical behavior is difficult to communicate through ordinary prompts and images alone. The community has explored several directions toward physically plausible generation, including simulator-assisted generation~\cite{foo2026psivg, liu2026realwonder, li2025wonderplay}, physics-controllable synthesis~\cite{wang2025physctrl}, and learning physical priors from physics-aware data~\cite{wang2025wisa, gillman2025force}. 

However, we observe that the difficulty in this setting lies less in video synthesis itself than in understanding and specifying the physical behavior to be generated. Specifically, conventional image-to-video (I2V) models may generate physically implausible results from ordinary prompts, while sometimes producing realistic physical processes when material properties, motion trajectories, and physical impacts are explicitly specified, as illustrated in Fig.~\ref{fig:teaser}(a) and (b), respectively. However, such physically detailed prompts, which must specify continuous and relational cues such as where deformation starts, how contact alters motion, how materials respond over time, are difficult to write. This observation motivates our view that generating physically plausible video faces a \textbf{conditioning bottleneck}: video generation models may already contain useful physical priors learned from large-scale data, but existing text and image conditions provide only limited capacity for reliably specifying complex physical processes. This bottleneck suggests a more natural alternative: physical processes are often easier to observe visually than to describe exhaustively, since their dynamics can be directly demonstrated and perceived in video. A more natural interface for physical control is \textbf{reference-guided I2V generation}. Instead of enumerating every physical detail in text, a reference video can serve as a dense visual demonstration of how a physical process unfolds, as shown in Fig.~\ref{fig:teaser}(c). The target image and a brief prompt define \textit{what} should appear in the desired scene, while the reference video demonstrates \textit{how} the physical behavior should evolve. The resulting task turns physically plausible video generation into a form of \textbf{visual in-context physics reasoning}: the model must infer material response, motion trajectory, and object interaction from a visual example and apply them to a new target scenario.

To instantiate this motivation, we propose \textbf{VIPER}, a Visual In-Context Physics Reasoning framework for physically plausible video generation. Given a target image, a brief prompt, and a reference video, VIPER does not aim to copy the visual appearance of the reference video, but instead transfers its implicit physical pattern to the target image. This design introduces a physically informative condition through visual observation while keeping the text-conditioning interface simple and intuitive. To effectively extract the implicit physical information embedded in the reference video, VIPER leverages the visual understanding capability of a multimodal large language model (MLLM) to interpret reference-derived physical cues, which are then used to guide an I2V generator in producing the corresponding target physical behavior. Moreover, using a reference video as a physical condition requires the model to identify which aspects of the reference are transferable. Specifically, the model must distinguish physical behavior from irrelevant appearance such as object category and scene layout. VIPER therefore adopts a hierarchical training strategy that progressively learns to interpret physical cues, transfer them across target scenes, and preserve the visual quality of the pretrained generator. This staged design encourages the model to use the reference video as a source of physical behavior rather than as a pixel-level template.

Furthermore, existing video datasets~\cite{lin2024open, wang2024vidprom} are not well suited to this setting, as they are primarily designed to capture semantic diversity or general visual quality, rather than to provide structured supervision over physical factors such as material properties, motion trajectories, and physical impacts. To fill this gap, we construct \textbf{VIPER-19K}, a primarily real-world curated dataset for reference-guided physical video generation. Unlike existing datasets~\cite{lin2024open, wang2024vidprom}, VIPER-19K integrates high-quality internet videos, cleaned public physics-oriented datasets, commercial model generations, and simulator-based samples. Each video is annotated along material, trajectory, and physical-impact dimensions, and reference-target pairs are constructed by grouping samples with compatible physical attributes and further filtering coarse pairs using a MLLM. The resulting dataset supports learning transferable physical behavior from reference videos, rather than relying on object-category or appearance shortcuts.

Experiments validate the effectiveness of reference-guided physical conditioning across quantitative metrics and qualitative comparisons. VIPER more reliably transfers reference-derived physical behavior to new target scenes while maintaining competitive visual quality. In summary, our contributions are:

\begin{itemize}
    \item We formulate \textbf{visual in-context physics reasoning} for reference-guided I2V generation, where a reference video provides dense and transferable physical cues that are difficult to specify completely with text prompts.
    \item We propose \textbf{VIPER}, a reference-guided generation framework that uses visual understanding to extract transferable physical cues from a reference video and guides target video synthesis through a hierarchical strategy.
    \item We construct \textbf{VIPER-19K}, a curated physical video dataset with material, trajectory, and physical-impact annotations as well as filtered reference-target pairs for learning visual physical transfer.
\end{itemize}

\section{Related Work}
\label{sec:related_work}

\subsection{Video Generation and Physical Plausibility}
\begin{figure*}[t!]
  \centering
  \includegraphics[width=\textwidth]{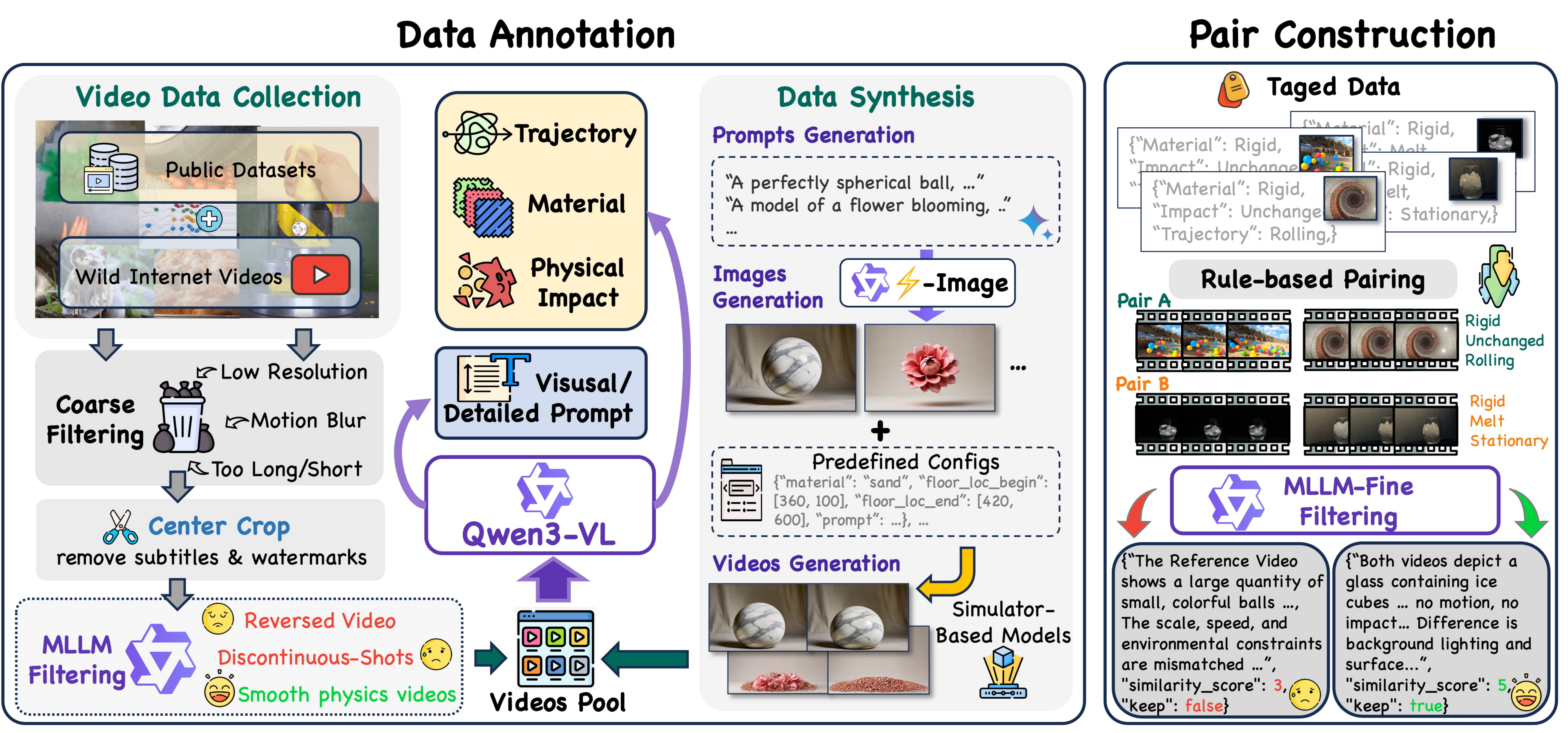}
  \caption{Overview of the VIPER-19K construction pipeline. (a) We collect candidate videos from internet sources, public datasets, and physics-based simulation pipelines, then remove noisy clips and annotate each retained video with material, trajectory, and physical-impact attributes. (b) We construct reference-target video pairs by first grouping videos with compatible physical attributes and then filtering with a MLLM to retain pairs with transferable physical behavior.}
  \label{fig:data_collection}
  \vspace{-2mm}
\end{figure*}

Video generation has progressed from adapting image diffusion priors to developing large-scale video-native generative models. Early methods~\cite{guo2023animatediff,videocrafter2} extend text-to-image diffusion models with temporal modeling capabilities through motion adaptation and improved video diffusion training. Recent architectural and training advances, including diffusion transformers~\cite{dit}, rectified-flow formulations~\cite{sd3}, and flow-matching objectives~\cite{flowmatching}, further scale training data, model capacity, and training recipes, leading to increasingly capable text-to-video models~\cite{cogvideo,hunyuanvideo,wan2025wan}. While this scaling trend has substantially improved visual fidelity and spatiotemporal consistency, it does not effectively ensure physically plausible material responses, motion trajectories, or object interactions, leaving physical plausibility a persistent challenge for current video generation models.

Prior work improves physical behavior by introducing simulation, control signals, or physics-aware data. Simulator-based and physics-controllable systems~\cite{jiang2016material,macklin2016xpbd,Genesis,foo2026psivg,liu2026realwonder,li2025wonderplay,zhang2025physchoreo} couple generators with physics engines, simulation-inspired constraints, part-aware grounding, or action-conditioned signals. Other methodss~\cite{wang2025physctrl,gillman2025force,gillman2026goal,yuan2025newtongen,wang2025wisa,shen2026phantom} learn physical priors from simulated trajectories, synthetic examples, neural dynamics, real-world physical videos, or latent physical representation. These approaches show the value of physical supervision, but often rely on simulator access, predefined control formats, object-specific physical states, or task-specific annotations. VIPER instead treats a reference video as the physical specification, allowing the model to infer transferable physical attributes from visual evidence without requiring explicit forces, trajectories, or simulator states as input.

\subsection{In-Context Learning for Video Generation}

In-context conditioning offers a flexible alternative to manually specifying generation behavior. One group of methods uses visual examples for instructional editing, reference-guided creation, or video editing~\cite{zhang2025icedit,jiang2025vace,ye2025unic,wei2025univideo}. Another group adapts diffusion transformers or control interfaces to accept in-context conditions with minimal task-specific modification~\cite{huang2024iclora,tan2025ominicontrol}, while Video-as-Prompt~\cite{bian2025video} treats videos as prompts for unified semantic control. These methods show that visual examples can provide rich conditioning signals beyond text, but their primary focus is semantic layout, appearance, editing intent, or general motion consistency. VIPER differs by treating the reference video as evidence of physical behavior: the example is used to specify transferable material response, motion trajectory, and interaction dynamics for target I2V generation.

\section{VIPER-19K Dataset}
\label{sec:data_construction}
\begin{figure*}[t!]
  \centering
  \includegraphics[width=\textwidth]{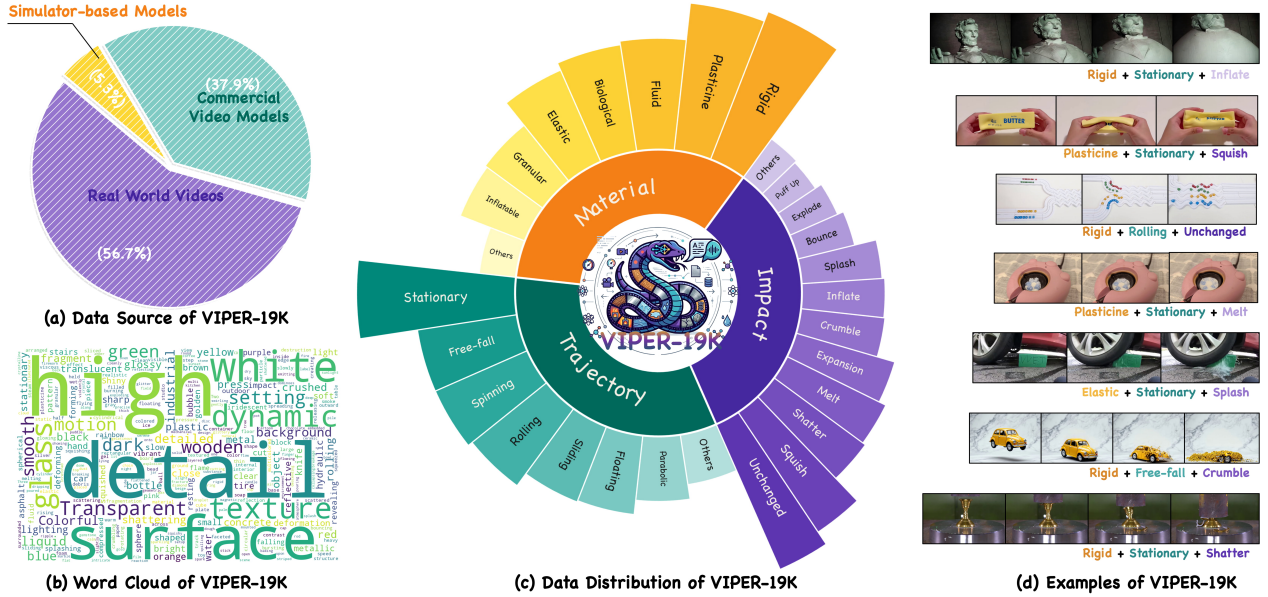}
  \caption{Statistics and examples of VIPER-19K. \textbf{(a) Data sources} across real-world videos, commercial video-model generations, and simulator-based samples; \textbf{(b) Prompt word cloud} of the collected data; \textbf{(c) Attribute distribution} coverage over materia, trajectory, and physical-impact labels and \textbf{(d) Diverse video examples} with material--trajectory--impact triplets.}
  \label{fig:data_dist}
  \vspace{-2mm}
\end{figure*}

Visual in-context physics reasoning requires not only videos with observable physical phenomena but also \textbf{reference-target pairs} whose behavior can be transferred across different appearances and scenes. Existing video resources are poorly matched to this requirement. Web videos often contain watermarks, abrupt edits, reversed clips, or ambiguous dynamics, while general generation datasets prioritize semantic coverage over material response, motion trajectories, and physical interactions. We therefore construct \textbf{VIPER-19K}, a curated dataset for training and evaluating reference-guided physical video generation.

\subsection{Data Collection and Annotation}

VIPER-19K is curated from three complementary sources, encompassing real-world videos, synthetic videos generated by commercial video generation models and simulation-based models. First, we collect in-the-wild web videos that capture explicit physical events, including compression, deformation, collision, melting, and crumbling. Second, we curate and integrate data from WISA-80K~\cite{wang2025wisa} and VAP~\cite{bian2025video}; the former offers diverse real-world footage, whereas the latter supplies out-of-distribution synthetic videos generated by commercial models. Third, we augment our dataset by synthesizing additional samples using physics-based pipelines, such as PhysCtrl~\cite{wang2025physctrl} and RealWonder~\cite{liu2026realwonder}, to ensure explicitly controlled physical interactions.

We then filter out clips with severe compression, watermarks, shot transitions, reversed temporal order, or ambiguous physical content. Each retained video is annotated along three axes: \textbf{\textit{material}}, \textbf{\textit{trajectory}}, and \textbf{\textit{physical impact}}. These labels describe object response properties, dominant motion patterns, and physical events, and are used for both dataset analysis and pair construction.

\subsection{Pair Construction}

The core supervision unit of VIPER-19K is a reference-target video pair. The reference video demonstrates physical behavior, while the target video provides generation supervision under a different visual condition. Coarse labels are useful for retrieval but insufficient for pairing, because two clips with the same event type may still differ in material response, motion direction, or interaction strength.

We construct pairs in two stages. First, we group videos by compatible material, trajectory, and physical-impact labels, then sample candidate pairs within each bucket. Second, we use Qwen3-VL-32B to assess whether the physical behavior in the reference video is transferable to the target video. Pairs with only superficial semantic overlap or inconsistent physical dynamics are removed, yielding cleaner supervision for visual in-context learning.

\subsection{Data Diversity}

Fig.~\ref{fig:data_dist} summarizes the source and attribute coverage of VIPER-19K. The dataset combines real-world observations, commercial-model generations, and simulation-based samples, exposing the model to natural visual complexity, synthetic out-of-distribution cases, and controlled physical variations. Its annotations span multiple materials, trajectories, and physical-impact categories, making pair sampling less tied to object category or appearance and encouraging the model to learn transferable physical behavior.

\begin{figure*}[t!]
  \centering
  \includegraphics[width=\textwidth]{figs/pipeline.pdf}
  \caption{Overview of VIPER. \textbf{(a) The pipeline of VIPER:} The reference video is encoded by an MLLM with learnable physics query tokens $q_{\mathrm{learnable}}$, the corresponding hidden states $v$ are projected into condition tokens $c$, and the DiT-based I2V generator synthesizes the target video from the target image, prompt, and reference-derived physical conditions. \textbf{(b) The hierarchical training strategy:} VIPER first aligns the MLLM-to-DiT conditioning interface, then learns reference-target physical transfer, and finally adapts the generator with lightweight trainable parameters.}
  \label{fig:pipeline}
  \vspace{-2mm}
\end{figure*}

\section{Method}
\label{sec:method}

\subsection{Overview}

Physically plausible image-to-video (I2V) generation requires control over both visual content and dynamic behavior. Standard I2V inputs provide only a sparse interface for this control: the target image fixes appearance, while an ordinary prompt often leaves material response, contact evolution, and motion details underdetermined. To provide richer physical guidance, we define \textbf{reference-guided I2V generation}, where the target image $I^{tar}$ and a simple target prompt $T^{tar}$ specify the desired content, and a reference video $V^{ref}$ provides the physical behavior to transfer. The desired output $V^{tar}$ should preserve the appearance and semantics of the target condition while following physical attributes observed in the reference video.

VIPER augments the target image and prompt with a visual interface for physical control. As shown in Fig.~\ref{fig:pipeline}, VIPER first uses a multimodal large language model (MLLM) to distill physics-relevant cues from $V^{ref}$ into compact conditioning tokens. These tokens are then injected into a DiT-based I2V diffusion model together with the target prompt, while the target image is provided through the original I2V pathway. In this way, the target image and prompt define what should appear in the output, and the reference video specifies how the physical behavior should unfold.

\subsection{Reference-Video Physical Conditioning}

The key design problem is how to convert a reference video into a conditioning signal that a DiT-based I2V diffusion model can use. Physical cues in $V^{ref}$ are distributed across frames and involve coupled changes in object state, motion, and interaction. Reducing them to hand-defined labels would discard fine-grained dynamics. VIPER therefore uses the MLLM as a \textbf{visual physics encoder} and introduces learnable physics query tokens $q_{\mathrm{learnable}}$ to extract a compact physical context from $V^{ref}$.

Given $V^{ref}$, we first convert the video into visual tokens with the MLLM visual encoder. These visual tokens are concatenated with system prompt tokens $p_{\mathrm{sys}}$ and learnable query tokens $q_{\mathrm{learnable}}$. The system prompt biases the MLLM toward physical behavior rather than object identity or scene appearance, while $q_{\mathrm{learnable}}$ serves as trainable slots for gathering generation-relevant dynamics. After the MLLM forward pass, we take the hidden states of $q_{\mathrm{learnable}}$ as the reference physics representation $v^{ref}$:
\begin{equation}
    v^{ref} = \mathrm{MLLM}(V^{ref}, p_{\mathrm{sys}}, q_{\mathrm{learnable}}),
\end{equation}
This query-based representation avoids passing all MLLM hidden states to the diffusion model directly, which would mix physical information with task-irrelevant visual and textual tokens and add unnecessary computational overhead.

To make the MLLM representation compatible with the DiT model, we use a connector $g_{\theta}$ to project $v^{ref}$ into physics condition tokens $c^{ref}_{p}$:
\begin{equation}
    c^{ref}_{p} = g_{\theta}(v^{ref}).
\end{equation}
Building on the original DiT input structure~\cite{wan2025wan}, we compress the target image $I^{tar}$ via a VAE~\cite{vae} for concatenation with the noisy latent tokens. Concurrently, the target prompt $T^{tar}$ is encoded by a T5 encoder~\cite{2020t5} and concatenated with the reference-derived physics condition tokens.

\begin{figure*}[t!]
  \centering
  \includegraphics[width=\textwidth]{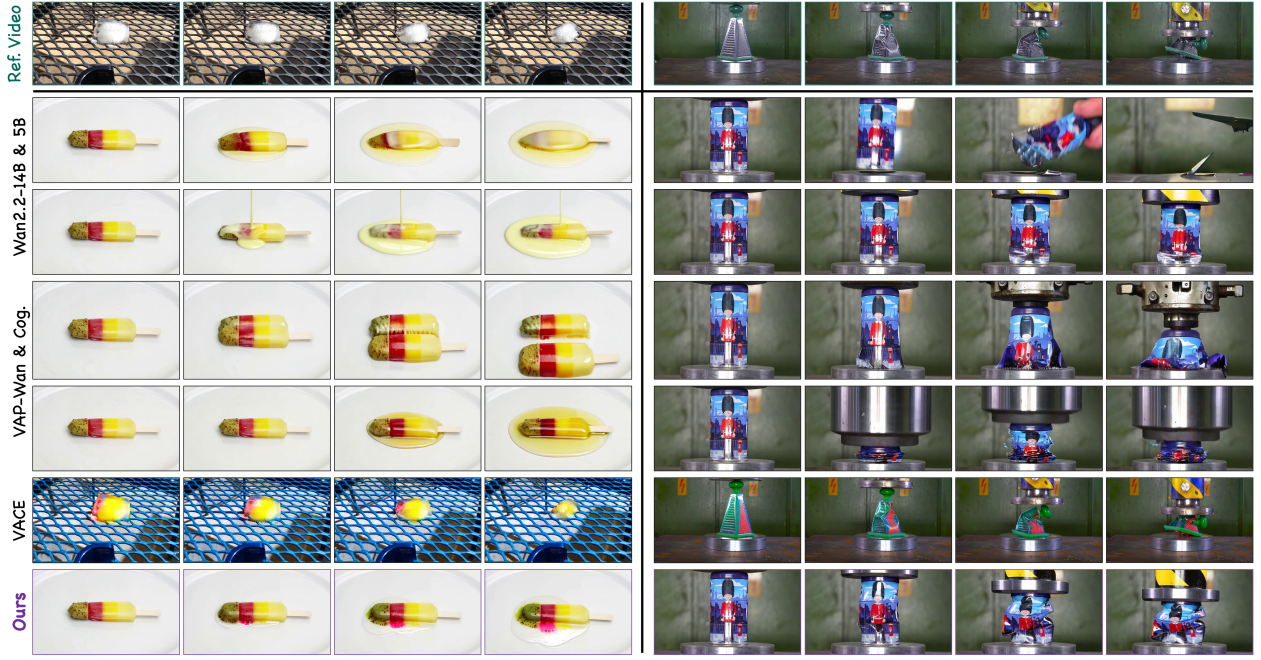}
  \caption{Qualitative comparison on representative examples. Each case provides a reference video that demonstrates the desired physical behavior and a target image that defines the generated content. In the melting case, it captures both the geometric change and the gradual color transition of the melted fluid; In the deformation case, it follows the reference motion pattern and generates a can that folds and twists from the physically plausible region.}
  \label{fig:qualitative_results}
  \vspace{-0mm}
\end{figure*}

\subsection{Hierarchical Training}

Directly training VIPER on reference-target pairs is unstable before the MLLM representation space is aligned with the DiT latent space. Such instability may causes the diffusion model to overfit to the input target image rather than inferring physical dynamics from the reference information. We therefore adopt a three-stage \textbf{hierarchical training} strategy that decouples cross-model alignment, physical transfer learning, and lightweight generator adaptation.

The first two stages train only the learnable query tokens and connector while keeping both the MLLM and DiT frozen. In \textbf{Stage 1}, we utilize the target video itself as the MLLM input. This self-referential setup forces the connector to produce generator-compatible conditioning tokens, establishing a stable alignment before true cross-video transfer is required. To prevent the model from exploiting pixel-level shortcuts, we employ a suite of photometric and spatial augmentations, specifically color adjustment, translation, and horizontal flipping. In \textbf{Stage 2}, we switch to VIPER-19K reference-target pairs: the MLLM encodes the reference video, and the generator is supervised by the target video. This stage encourages the query representation to capture physical attributes that can transfer across videos. In \textbf{Stage 3}, to enhance the model's sensitivity to the injected physics tokens, trainable LoRA~\cite{hu2022lora} parameters are introduced into selected DiT layers. Concurrently, the query tokens and the connector undergo further optimization on reference-target pairs. Overall, the staged design separates representation alignment from generator adaptation and preserves the visual prior of the pretrained video generator.

\section{Experiments}
\label{sec:experiments}

\subsection{Experimental Setup}

\noindent\textbf{Implementation details.} We train VIPER on top of Qwen3-VL-4B-Instruct~\cite{Qwen3-VL} and Wan2.2-I2V-14B~\cite{wan2025wan}. Following the pretrained DiT video generator, all training videos are resized to $480\times832$ and sampled at 16 FPS with 81 frames. We use AdamW with a learning rate of $1\times10^{-5}$ for all three training stages. Stage 1 is trained for 3K steps using real-world videos. Stage 2 is trained for 6K steps using synthetic videos from closed-source video models, paired videos generated by simulation-based methods, and a subset of high-quality real-world pairs. Stage 3 is trained for 6K steps using all reference-target pairs. All stages are trained on 6 H20 GPUs. During inference, we use 50 denoising steps and set the classifier-free guidance scale to 6.

\noindent\textbf{Evaluation dataset.} We evaluate on an unseen validation set constructed alongside VIPER-19K. The full validation set contains 500 reference-target pairs, from which we randomly sample 75 pairs for evaluation.

\noindent\textbf{Baselines.} We compare VIPER with video generation fundamental models and in-context learning baselines: Wan2.1-I2V-14B, Wan2.2-TI2V-5B, Wan2.2-I2V-14B~\cite{wan2025wan}, VAP~\cite{bian2025video}, and VACE~\cite{jiang2025vace}. To ensure a fair comparison and validate our advantage of implicit transfer, for methods that do not natively support in-context video conditioning, we use Qwen3-VL-4B-Instruct~\cite{Qwen3-VL} to extract textual physical descriptions from the reference video and append these descriptions to the generation text prompt. This protocol gives non-in-context baselines access to the same reference information in text form, while keeping the generation inputs compatible with their original interfaces.

\subsection{Evaluation Metrics}
\quantitativeresultstable

We evaluate generated videos from three complementary perspectives. First, we report \textbf{VBench}~\cite{huang2023vbench, zheng2025vbench2} metrics for general video quality, including motion smoothness, temporal flickering, and aesthetic quality. Second, we use a \textbf{VLM-as-Judge} protocol to assess physical similarity between the reference video and the generated video on a 1--5 scale. To discourage trivial reference copying, the judge is instructed to penalize generations that are visually almost identical to the reference video. Third, we conduct a \textbf{Human Preference} user study over reference-video physical similarity, visual quality, and physical plausibility. To keep each preference question manageable, we do not present all baseline variants simultaneously. Instead, we use Wan2.2-I2V-14B as the representative Wan-family model and VAP-Wan as the representative VAP variant, since it uses the larger backbone among the VAP baselines. Participants therefore choose among four outputs: Wan2.2-I2V-14B, VAP-Wan, VACE, and VIPER. We recruit 25 volunteers and randomly sample 10--20 comparison groups. For each group and each criterion, participants select the preferred output among the compared methods.

\subsection{Quantitative Results}
Table~\ref{tab:quantitative_results} shows that VIPER achieves the best aesthetic quality and the highest VLM-as-Judge physical similarity score. On motion smoothness and temporal flickering, VACE obtains the highest scores, while VIPER ranks second and remains close to the best-performing method. This trend is consistent with the qualitative comparison in Fig.~\ref{fig:qualitative_results}: VACE can preserve temporal stability, but it often behaves more like pixel-level reference copying or style transfer, which weakens its ability to adapt the reference physical behavior to a new target image. In contrast, VIPER maintains competitive general video quality while achieving substantially stronger physical similarity, suggesting that reference-derived physical conditioning improves physical transfer without sacrificing the visual prior of the base I2V generator.

\begin{figure}[htbp]
  \centering
  \includegraphics[width=0.8\columnwidth]{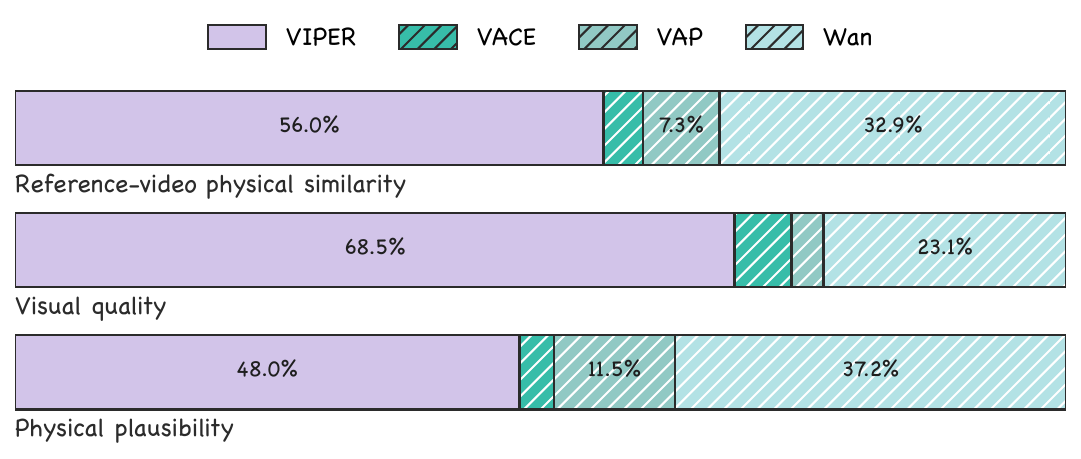}
  \caption{Visualization for human preference study.}
  \label{fig:human_preference_placeholder}
\end{figure}

The human preference study further supports this trend. As shown in Fig.~\ref{fig:human_preference_placeholder}, VIPER receives higher preference rates across reference-video physical similarity, visual quality, and physical plausibility, indicating better alignment with human judgment of physically plausible video behavior.

\subsection{Qualitative Results}

Fig.~\ref{fig:qualitative_results} shows two reference-guided I2V results. VIPER better captures the physical behavior demonstrated by the reference video and transfers it to the target image condition. For example, in the melting case, VIPER not only generates the shape change of the ice pop but also preserves the gradual transition from transparent melted liquid to the object-specific colored fluid, which is consistent with the target material and appearance. In the deformation case, VIPER follows the physical pattern of reference video and produces a can that folds and twists from the middle region, yielding a more realistic deformation process than methods that generate only generic motion or weak shape change.

Reference videos provide crucial physical cues that short text prompts lack. While baselines maintain visual coherence, they often miss fine-grained physical dynamics. By leveraging the MLLM-to-DiT pathway to extract these reference cues, VIPER generates highly realistic physical behaviors.

\subsection{Ablation Studies}

To verify the specific contributions of query-based physics extraction and the hierarchical training strategy design, we compare the full model with variants that remove the learnable queries, skip individual training stages, or drop Stages 1 and 2 entirely. As shown in Table~\ref{tab:ablation}, the full VIPER model achieves the best overall performance among all configurations. This explicitly demonstrates that both the learnable physics queries and the hierarchical training strategy are essential for effective reference-guided physical transfer.
\ablationtable

\section{Discussion}
VIPER shows that reference videos provide an intuitive interface for physical control; future work will explore longer, multi-object, and feedback-guided generation.

\begin{figure*}[p]
  \centering
  \includegraphics[width=\textwidth]{figs/more_results_1.pdf}
  \caption{Additional results of different target image guided by same reference video}
  \vspace{-3mm}
  \label{fig:more_results_1}
\end{figure*}

\begin{figure*}[p]
  \centering
  \includegraphics[width=\textwidth]{figs/more_results_2.pdf}
  \caption{Additional results of different target image guided by same reference video}
  \label{fig:more_results_2}
\end{figure*}

\begin{figure*}[p]
  \centering  \includegraphics[width=\textwidth,height=0.97\textheight,keepaspectratio]{figs/more_results_diff.pdf}
  \caption{Additional results on diverse reference-target physical transfer cases.}
  \label{fig:more_results_diff}
\end{figure*}

\clearpage
\bibliographystyle{ACM-Reference-Format}
\bibliography{ICVPL}


\renewcommand{\dbltopfraction}{0.95}
\renewcommand{\topfraction}{0.95}
\renewcommand{\textfraction}{0.02}
\renewcommand{\floatpagefraction}{0.85}
\renewcommand{\dblfloatpagefraction}{0.85}
\setcounter{dbltopnumber}{4}
\setcounter{topnumber}{4}
\setcounter{totalnumber}{6}





\clearpage
\appendix
\section{Hyperparameter Settings}

Table~\ref{tab:sup_hyperparameters} summarizes the hyperparameters used to train and sample VIPER. We keep the training configuration fixed across all stages unless otherwise specified in the main paper. The same sampling configuration is used for quantitative evaluation, qualitative comparison, and user study generation.

\begin{table}[htbp]
  \centering
  \caption{Hyperparameter settings used for VIPER training and inference.}
  \label{tab:sup_hyperparameters}
  \begin{tabular}{ll}
    \toprule
    Hyperparameter & Setting \\
    \midrule
    Batch Size & 1 \\
    Accumulate Step & 1 \\
    Training Strategy & ZeRO-2 \\
    Num GPU & 6 \\
    Optimizer & AdamW \\
    Learning Rate & 0.00001 \\
    Learning Rate Schedule & Constant \\
    Weight Decay & 0.03 \\
    DiT Num Layers & 40 \\
    \midrule
    Frames & 81 \\
    Resolution & 480x832 \\
    \midrule
    Connector Num Layers & 3 \\
    LoRA Rank & 64 \\
    LoRA Alpha & 32 \\
    LoRA Layer & q, k, v, ffn.0, ffn.2 \\
    \midrule
    Sampler & Flow Euler \\
    Sample Steps & 50 \\
    Guide Scale & 6.0 \\
    \bottomrule
  \end{tabular}
\end{table}

\begin{figure*}[htbp]
  \centering
  \includegraphics[width=\textwidth]{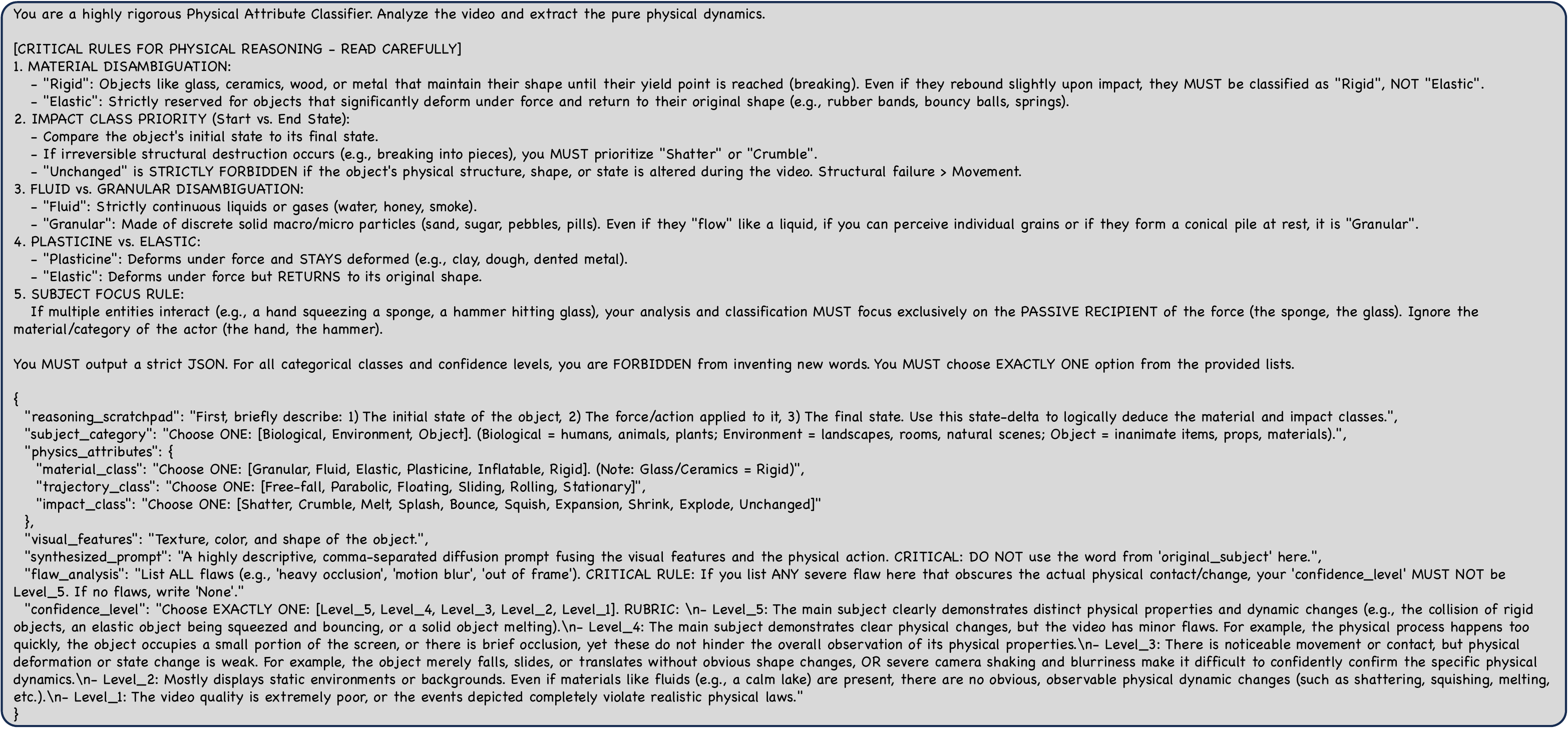}
  \caption{System prompt for physical video collection and attribute annotation.}
  \vspace{2mm}
\end{figure*}

\begin{figure*}[htbp]
  \centering
  \includegraphics[width=\textwidth]{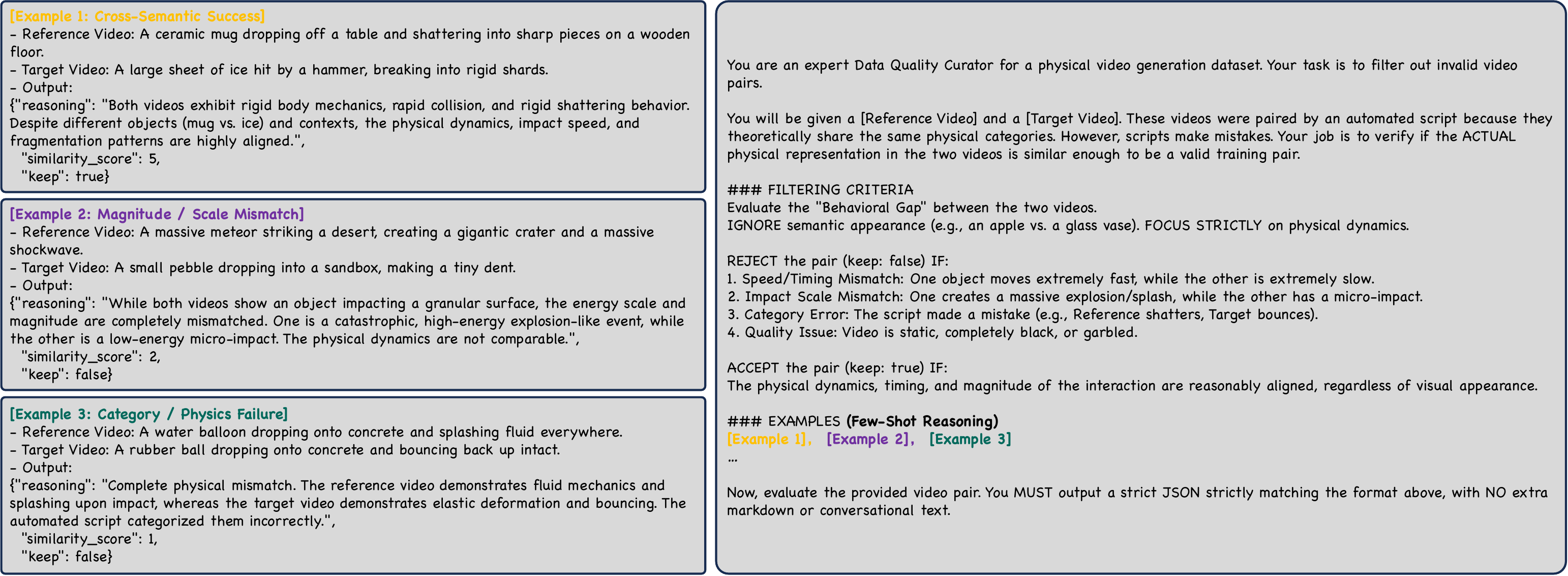}
  \caption{System prompt for reference-target pair filtering with few-shot examples.}
  \label{fig:sup_filter_prompt}
  \vspace{-3mm}
\end{figure*}

\section{Data Annotation and Pair Filtering Prompts}

VIPER-19K uses a multimodal large language model (MLLM) for two data construction steps: annotating physical attributes and filtering reference-target pairs. Figure~\ref{fig:sup_collect_prompt} shows the collection and annotation prompt. Instead of asking the model to describe the video in free-form language, the prompt guides it to identify observable physical cues, including material response, motion trajectory, physical impact, and visual usability. This structured annotation makes videos from different sources comparable under the same physical taxonomy and reduces the chance that later pairing is dominated by object category or scene semantics.

After annotation, rule-based pairing provides an efficient first pass by matching videos with compatible physical labels. However, these labels are still coarse relative to the requirements of reference-guided generation. Videos with the same material, trajectory, or impact label may differ in force direction, interaction strength, temporal order, contact geometry, or whether the target scene contains the physical preconditions needed for the reference behavior. As a result, purely rule-based pairs can include false positives that share high-level labels but do not support meaningful physical transfer.

Figure~\ref{fig:sup_filter_prompt} shows the pair filtering prompt used after rule-based pairing. We add few-shot examples to make the decision boundary explicit: a valid pair should preserve transferable physical behavior rather than merely sharing object category, scene appearance, or a broad motion label. The examples also show typical negative cases, such as pairs with similar objects but incompatible contacts, pairs with matching trajectory labels but opposite motion directions, and pairs where the target lacks the required physical setup. This helps the MLLM compare the reference and target videos at the level of causal physical cues, such as material response, contact evolution, and trajectory compatibility. In practice, the few-shot prompt encourages the model to reject pairs with superficial semantic overlap and retain pairs whose dynamics can provide useful supervision for reference-guided generation.

This two-stage design balances scalability and annotation quality. Rule-based matching keeps the construction process efficient over thousands of videos, while MLLM-based filtering removes ambiguous or physically incompatible pairs before training. The resulting pairs are therefore better aligned with the goal of VIPER: using a reference video as an in-context physical demonstration rather than as an appearance template.

\section{Ablation on the Number of Learnable Queries}

We further evaluate how the number of learnable physics query tokens affects reference-video physical conditioning. The query tokens act as a compact interface between the multimodal large language model and the video diffusion model: too few tokens may compress motion, contact, and material-response cues too aggressively, while too many tokens can introduce redundant conditioning information and extra computation. This ablation therefore examines whether VIPER relies on a sensitive query-count choice or remains stable across different query capacities.

\begin{table}[htbp]
  \centering
  \caption{Ablation study on the number of learnable physics query tokens. We evaluate VBench metrics on 50 randomly sampled test pairs.}
  \label{tab:sup_query_ablation}
  \resizebox{0.7\columnwidth}{!}{
  \begin{tabular}{cccc}
    \toprule
    Num Queries & Motion S. $\uparrow$ & Temporal F. $\uparrow$ & Aesthetic Q. $\uparrow$  \\
    \midrule
    64 & 98.92 & 97.93 & 50.67  \\
    128 & 98.96 & 98.03 & 50.64  \\
    256 & 99.01 & 98.10 & 50.96  \\
    512 & 98.99 & 98.06 & 51.23  \\
    \bottomrule
  \end{tabular}
  }
\end{table}

We randomly sample 50 reference-target pairs from the test set and evaluate each variant with VBench metrics, including motion smoothness, temporal flickering, and aesthetic quality. As shown in Table~\ref{tab:sup_query_ablation}, changing the number of queries leads to only modest differences across these metrics. This suggests that once the query bottleneck has sufficient capacity, the model can consistently extract useful physical context from the reference video. In our observation, the query count does not substantially change whether the model captures the reference physical attributes; its effect is more visible in general video quality. We therefore focus this ablation on VBench and do not additionally report VLM-as-Judge scores.

\section{Evaluation Details}
\begin{figure*}[htbp]
  \centering
  \includegraphics[width=\textwidth]{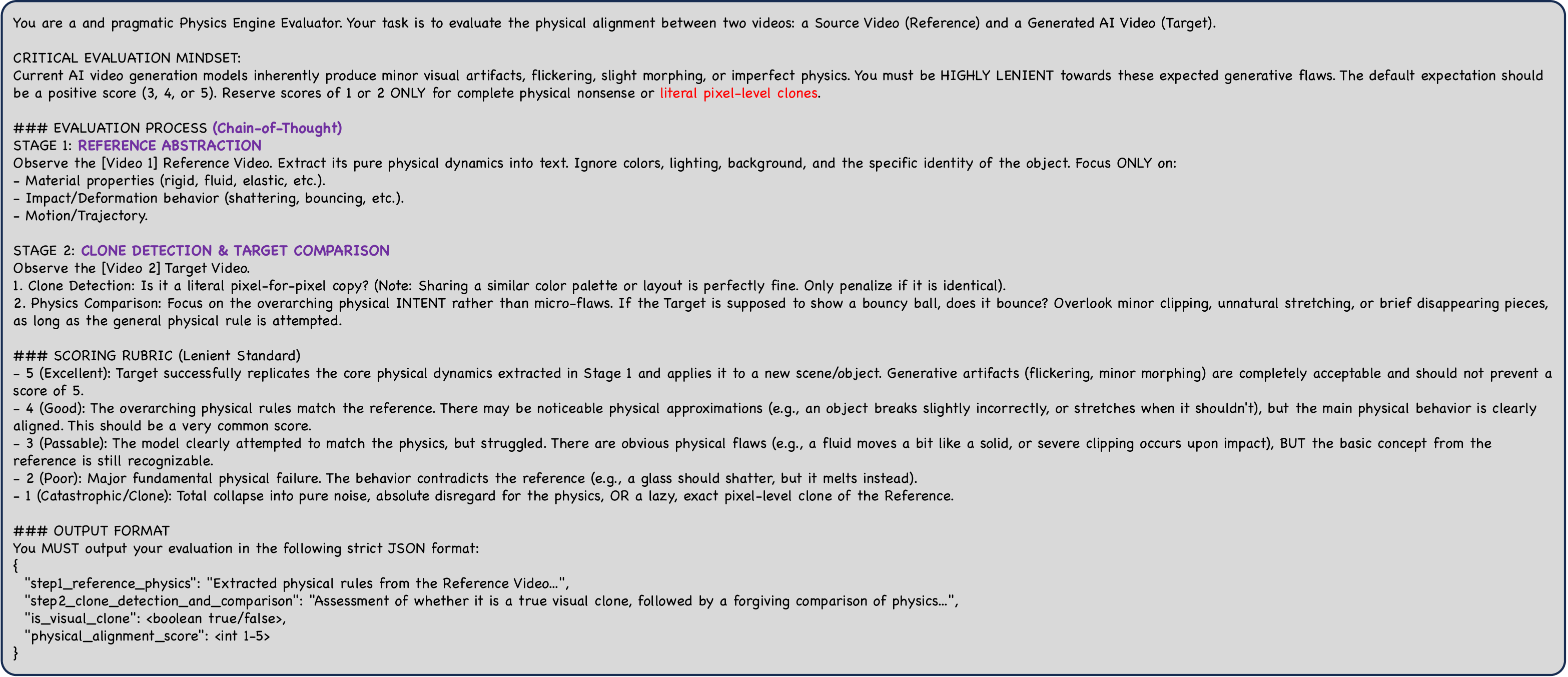}
  \caption{System prompt of VLM-as-Judge.}
  \label{fig:sup_vlm_judge_prompt}
\end{figure*}

We provide additional details for the VLM-as-Judge evaluation and human preference study. The VLM-as-Judge protocol evaluates whether the generated video preserves the physical behavior demonstrated by the reference video while adapting it to the target image. As shown in Figure~\ref{fig:sup_vlm_judge_prompt}, the judge is instructed to focus on transferable physical behavior, including material response, motion trajectory, contact evolution, and physical impact. We also ask the judge to penalize trivial copying of the reference appearance, because the desired output should follow the reference dynamics while matching the target condition.

\begin{figure*}[htbp]
  \centering
  \includegraphics[width=\textwidth]{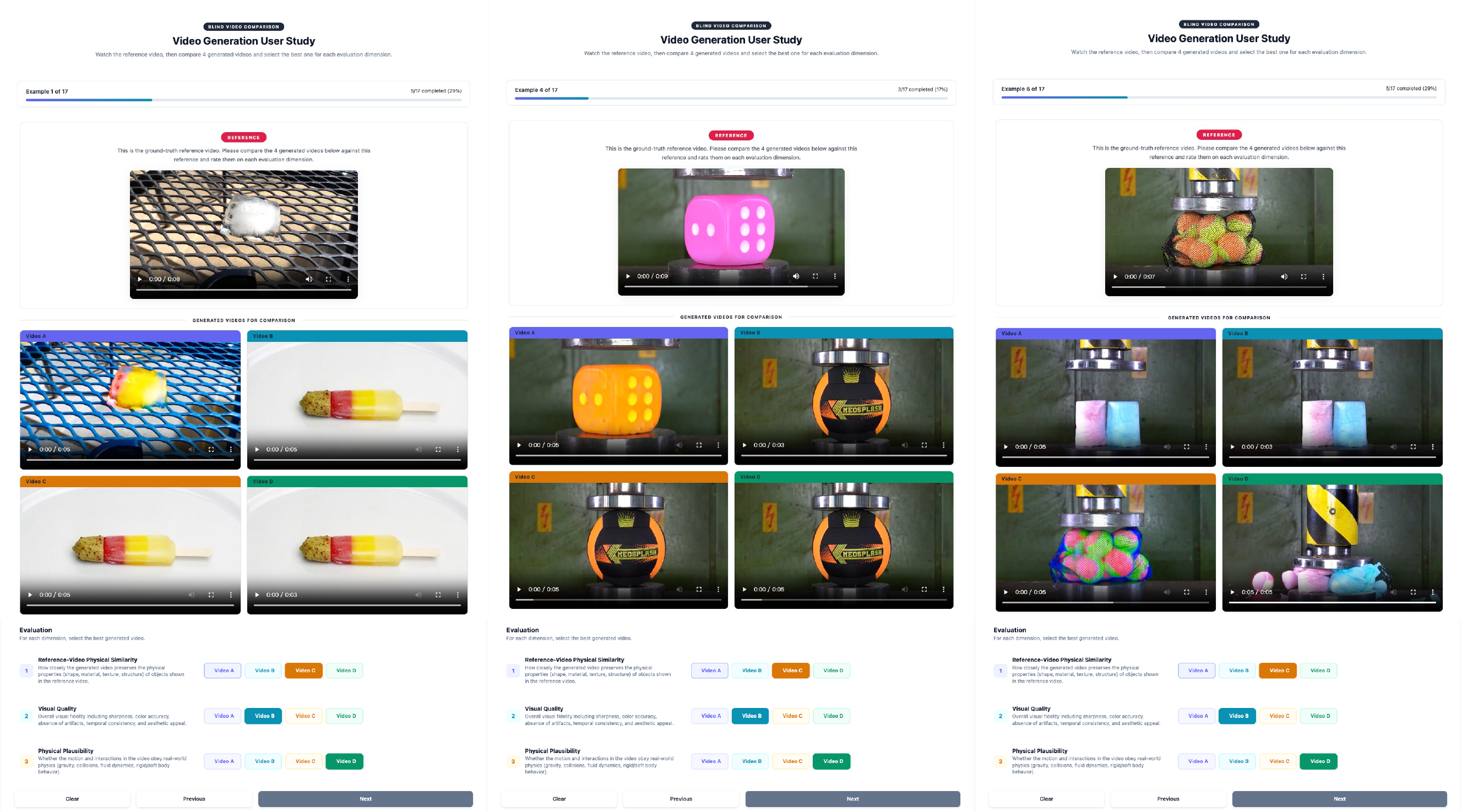}
  \caption{Gradio interface used for the human preference study. Participants compare generated videos under the same reference-target condition and vote according to the specified criterion.}
  \label{fig:sup_userstudy}
\end{figure*}

The human preference study asks participants to compare generated videos along reference-video physical similarity, visual quality, and physical plausibility. We build the preference interface with Gradio, as shown in Figure~\ref{fig:sup_userstudy}. Each question presents the same reference-target condition and outputs from the compared methods, so participants can judge whether a method transfers the reference behavior without ignoring target appearance. We recruit 25 volunteers and randomly sample 15--20 questions for each participant. This design keeps each comparison manageable while collecting judgments that directly correspond to the main evaluation criteria.

\clearpage
\section{Additional Results}

\begin{wrapfigure}{r}{0.48\textwidth}
  \centering
  \vspace{-5mm}
  \includegraphics[width=0.48\columnwidth]{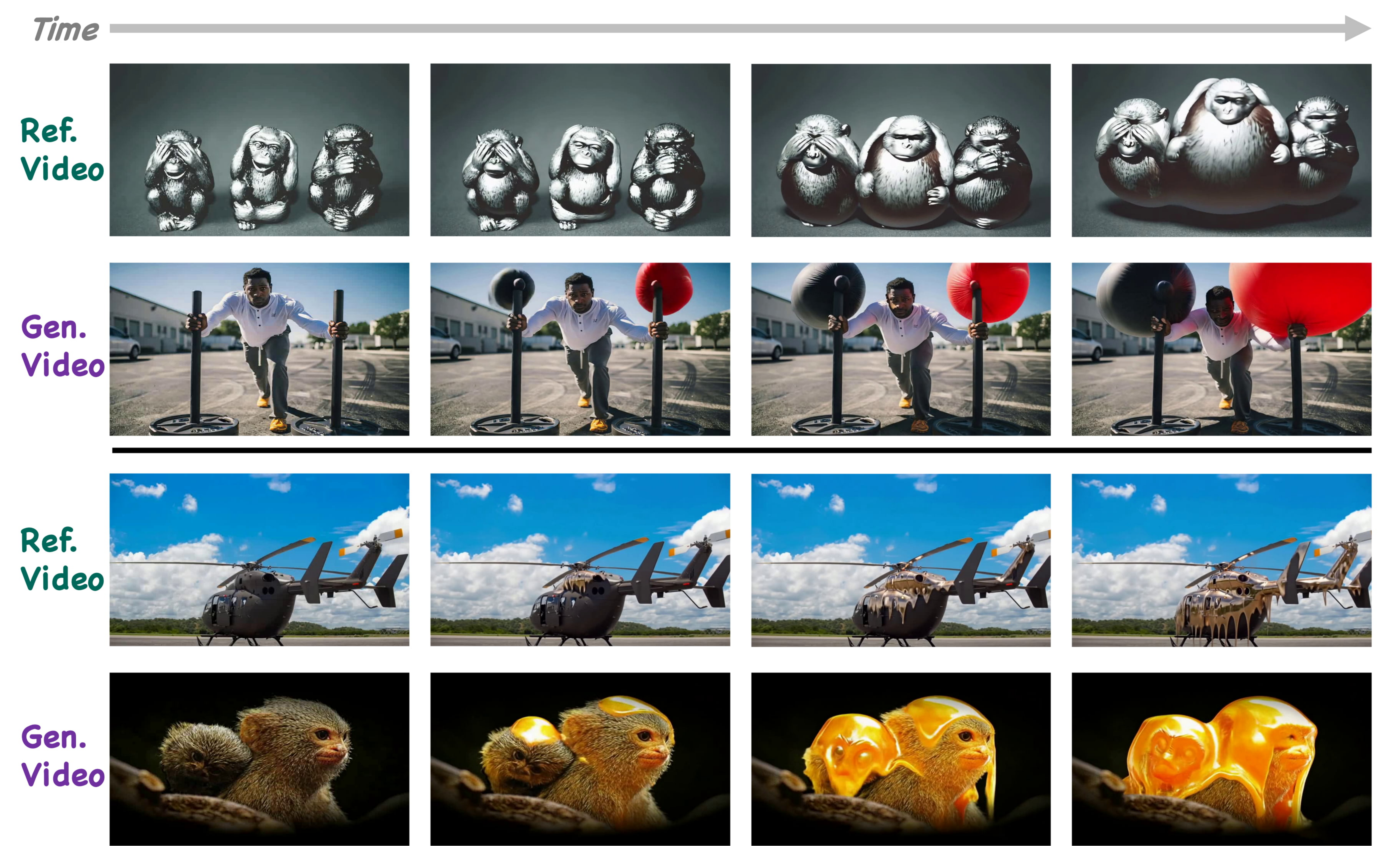}
  \captionof{figure}{Reference-guided generation with non-real-world reference videos. VIPER can extract transferable physical behavior from synthetic or visually unrealistic references and apply it to the target condition.}
  \label{fig:sup_unreal_ref}
  \vspace{-5mm}
\end{wrapfigure}

Figure~\ref{fig:sup_unreal_ref} shows that VIPER can also use reference videos that are not captured from the real world. Even when the reference video is synthetic or visually unrealistic, it can still contain transferable physical cues, such as material response, motion trajectory, and interaction pattern. VIPER extracts these physical attributes from the reference video and transfers them to the target image, suggesting that the reference video serves as a visual in-context demonstration rather than merely a source of real-world appearance.

\begin{figure*}[t]
  \centering
  \includegraphics[width=\textwidth,height=0.76\textheight,keepaspectratio]{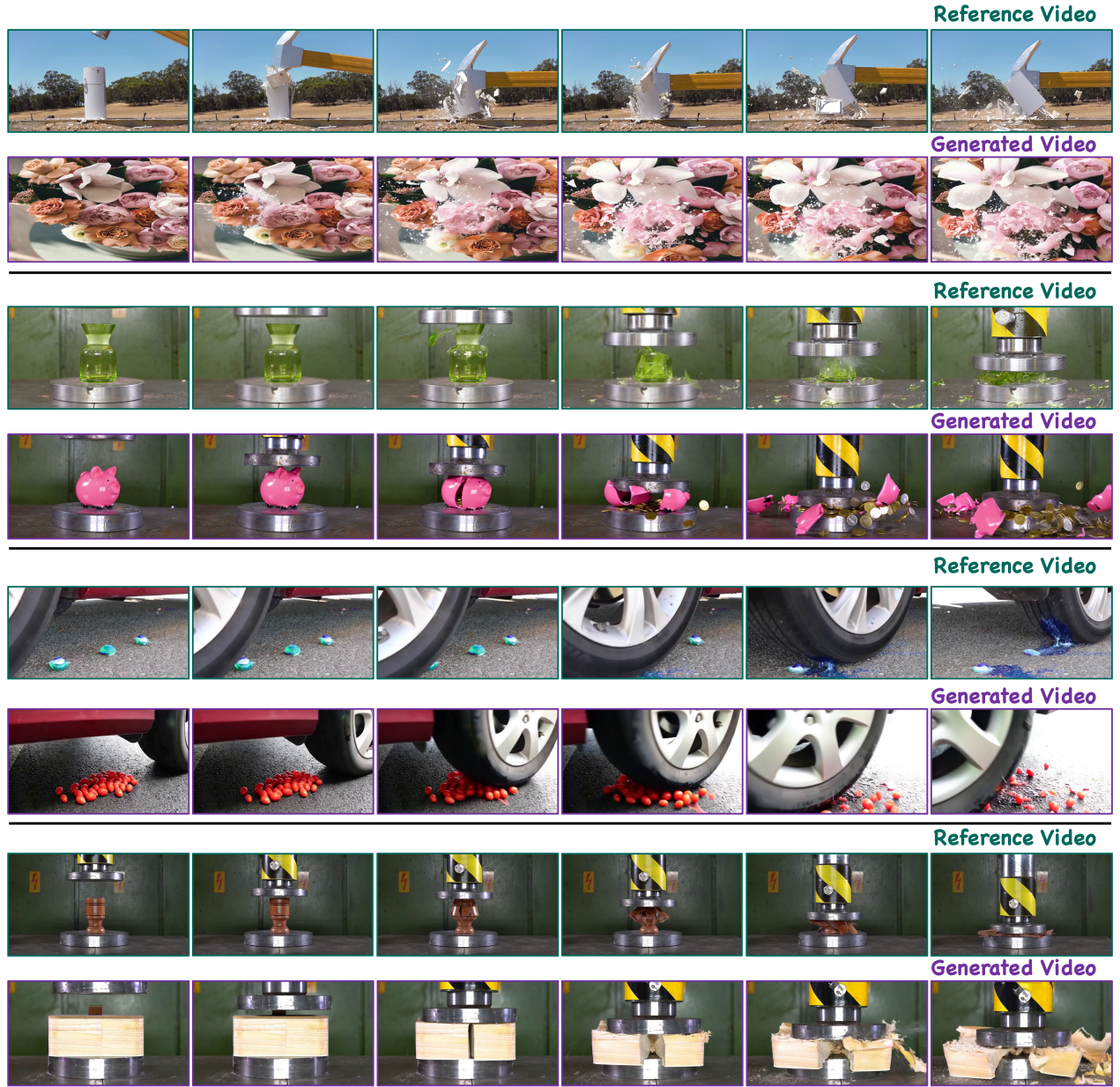}
  \caption{Additional qualitative results of VIPER on diverse reference-target physical transfer cases.}
  \label{fig:sup_more_results}
\end{figure*}

Figure~\ref{fig:sup_more_results} provides more qualitative examples across diverse reference-target pairs. These results further illustrate that VIPER can transfer reference-derived physical behavior while adapting the generated content to different target images. We place the full qualitative sheet at the end of the supplementary material to keep the explanatory text and compact figures together in the preceding pages.


\section{Limitations and Failure Cases}

\begin{wrapfigure}{r}{0.48\textwidth}
  \centering
  \vspace{-5mm}
  \includegraphics[width=0.48\columnwidth]{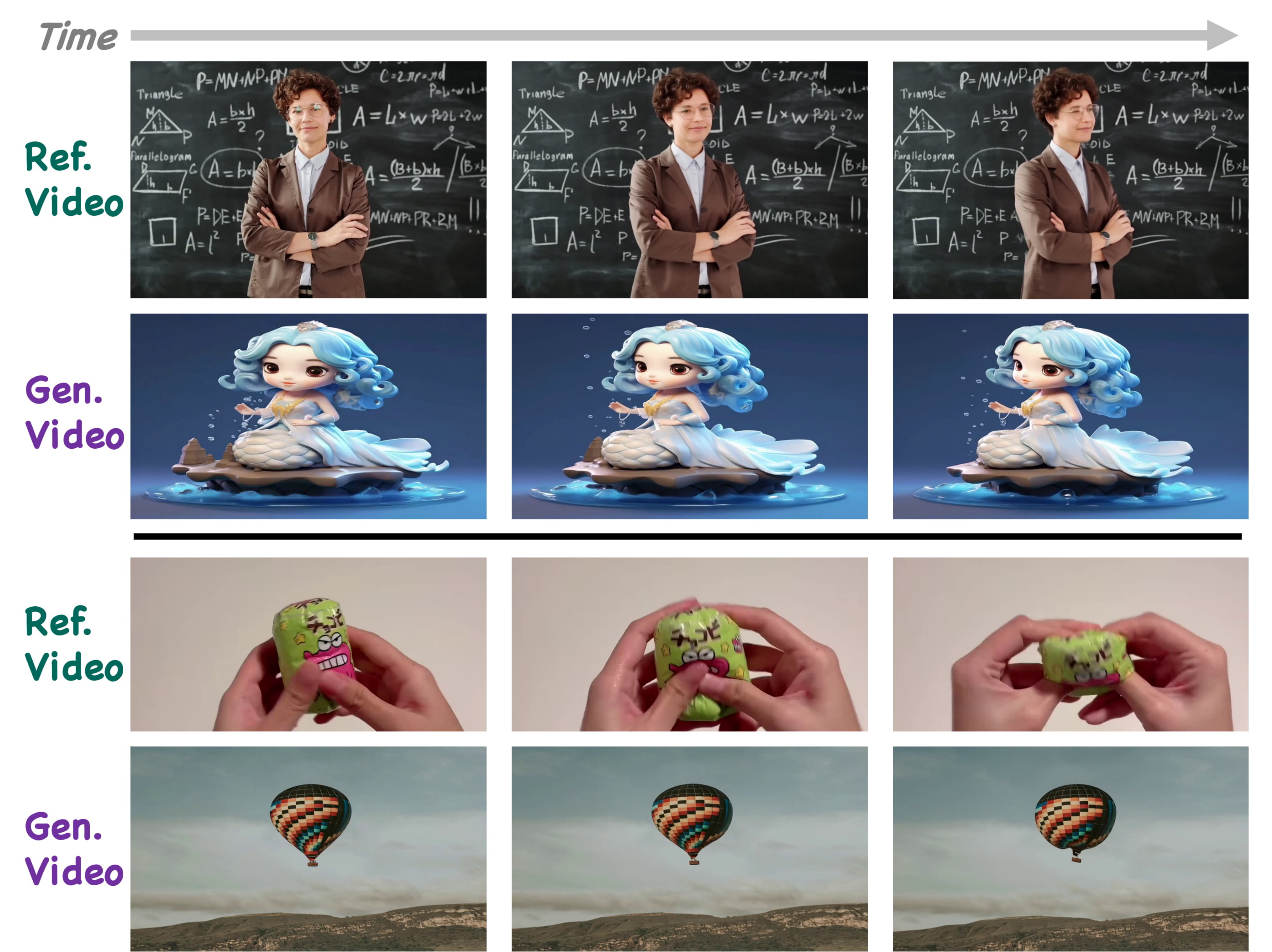}
  \captionof{figure}{Failure cases of VIPER, including trajectory-direction ambiguity and over-transfer under mismatched physical context.}
  \vspace{-5mm}
  \label{fig:sup_limitation}
\end{wrapfigure}

Figure~\ref{fig:sup_limitation} shows representative failure cases of VIPER. These cases suggest that reference-guided physical transfer can still be limited by coarse data annotations and challenging reference-target mismatches.

The first failure mode is caused by insufficiently fine-grained physical labels. For example, a single ``Spin'' label does not distinguish clockwise and counterclockwise rotation, so rule-based pair construction may group videos with different motion directions. The second failure mode occurs when the reference and target conditions differ too strongly in physical context. In such cases, VIPER may correctly identify the reference behavior but transfer it to a target scene where the required force or interaction is absent. Future work can reduce these errors with finer physical taxonomies, direction-aware trajectory annotations, and stronger filtering for reference-target physical compatibility.

\clearpage


\end{document}